\documentclass[letterpaper, 10 pt, conference]{ieeeconf}  
\IEEEoverridecommandlockouts  
\usepackage{float}
\usepackage{amsmath}
\usepackage{amssymb}
\usepackage{comment}
\usepackage{algorithm}
\usepackage{algpseudocode}
\usepackage{graphicx}
\usepackage{multirow}
\usepackage{subcaption}
\usepackage{array}
\usepackage{booktabs}
\usepackage{caption}
\usepackage{tabularx}
\usepackage{xcolor}

\usepackage{hyperref}
\newcommand{\projectpage}{\href{https://structured-diffuser.github.io}{\textcolor{black}{https://structured-diffuser.github.io}}}
\usepackage{threeparttable}

\newcolumntype{C}[1]{>{\centering\arraybackslash}p{#1}}

\newtheorem{remark}{Remark}

\begin{document}

\title{\LARGE \bf \textcolor{black}{Structured-Diffuser: 
Diffusion with Task-Conditioned Structured Priors for Motion Planning }  }

\author{
Amelie~Minji~Kim, Junling~Mei, Anqi~Wu, and Ye~Zhao%
\thanks{The authors are with the Institute for Robotics and Intelligent Machines,
the George W. Woodruff School of Mechanical Engineering, and the School of
Computational Science and Engineering, Georgia Institute of Technology,
Atlanta, GA, USA.
\texttt{\{mkim955,jmei66,anqiwu\}@gatech.edu, ye.zhao@me.gatech.edu}.}%
}
\maketitle

\IEEEpeerreviewmaketitle

\begin{abstract}
\textcolor{black}{We propose Structured-Diffuser, a diffusion planner that embeds task and motion structure directly into the noise model.} 
Unlike standard diffusion-based planners that rely on zero-mean, isotropic Gaussian corruption, we introduce task-conditioned structured Gaussians whose means and covariances are derived from Gaussian Process Motion Planning (GPMP), explicitly encoding trajectory smoothness and task semantics in the prior. 
We first formulate diffusion under a task-conditioned, non-isotropic Gaussian prior with closed-form forward and posterior expressions. Building on this formulation, our hierarchical design separates prior instantiation from trajectory denoising. 
At the upper level, sparse task-centric key states and timings are obtained, which instantiate a structured Gaussian prior (mean and covariance). At the lower level, the full trajectory is denoised under this prior, treating the upper-level outputs as noisy observations. 
\textcolor{black}{
Experiments across three motion-planning tasks show improved task success and training efficiency, with additional gains in data efficiency, trajectory smoothness, and position–velocity consistency where evaluated. 
} Ablation studies further show that explicitly structuring the corruption process provides benefits beyond neurally conditioning the denoising network alone. 
Overall, our approach concentrates the prior’s probability mass around task-relevant, temporally structured trajectories.  
\textcolor{black}{
We additionally demonstrate deployment on a physical G1 humanoid.}
Project page: \projectpage.
\end{abstract}

\section{Introduction}

Diffusion models have garnered increasing attention in robotic motion planning and shown strong flexibility across tasks \cite{carvalho2023mdp,janner2022planning,chi2023diffusion,ajay2022conditional}. 
However, most existing approaches adopt a default corruption process using zero-mean, isotropic Gaussian noise, overlooking the inherent structure of robotic trajectories. In contrast to image or text data,
{\color{black} robotic motion possesses intrinsic properties such as temporal smoothness (e.g., consistent evolution across time steps for dynamic feasibility) and can also be shaped by structured external information, such as task-imposed conditions (e.g., start and goal states or phase-specific constraints). 
These structured sources of information are typically incorporated through conditioning or guidance, rather than by shaping the corruption process itself \cite{carvalho2023mdp,chi2023diffusion,yan2025m2diffuser,mizuta2024cobl,pan2024model}. 

In this work, we propose to encode such structure in the form of a task-aware noise model. This design enables our diffusion planner to encode smoothness and task constraints directly in the noise prior. We realize this idea with a two-level hierarchical diffusion framework.} The upper-level planner produces sparse, task-centric key states (or key constraints).
We treat these key states as soft observations with explicit uncertainty, allowing imperfect predictions to guide sampling without hard constraints or rejection. 
The lower-level diffusion then generates the full trajectory under a task-conditioned Gaussian, obtained by conditioning a linear time-varying Gaussian process (LTV-GP) motion prior on the key states, inspired by Gaussian process motion planning (GPMP) \cite{mukadam2018continuous}. This yields a temporally correlated Gaussian whose mean is anchored by task-relevant key states and whose covariance encodes smooth motion between them. Accordingly, this framework produces temporally consistent and constraint-aware plans.

\begin{figure}[t]
\centering
\includegraphics[width=1\linewidth]{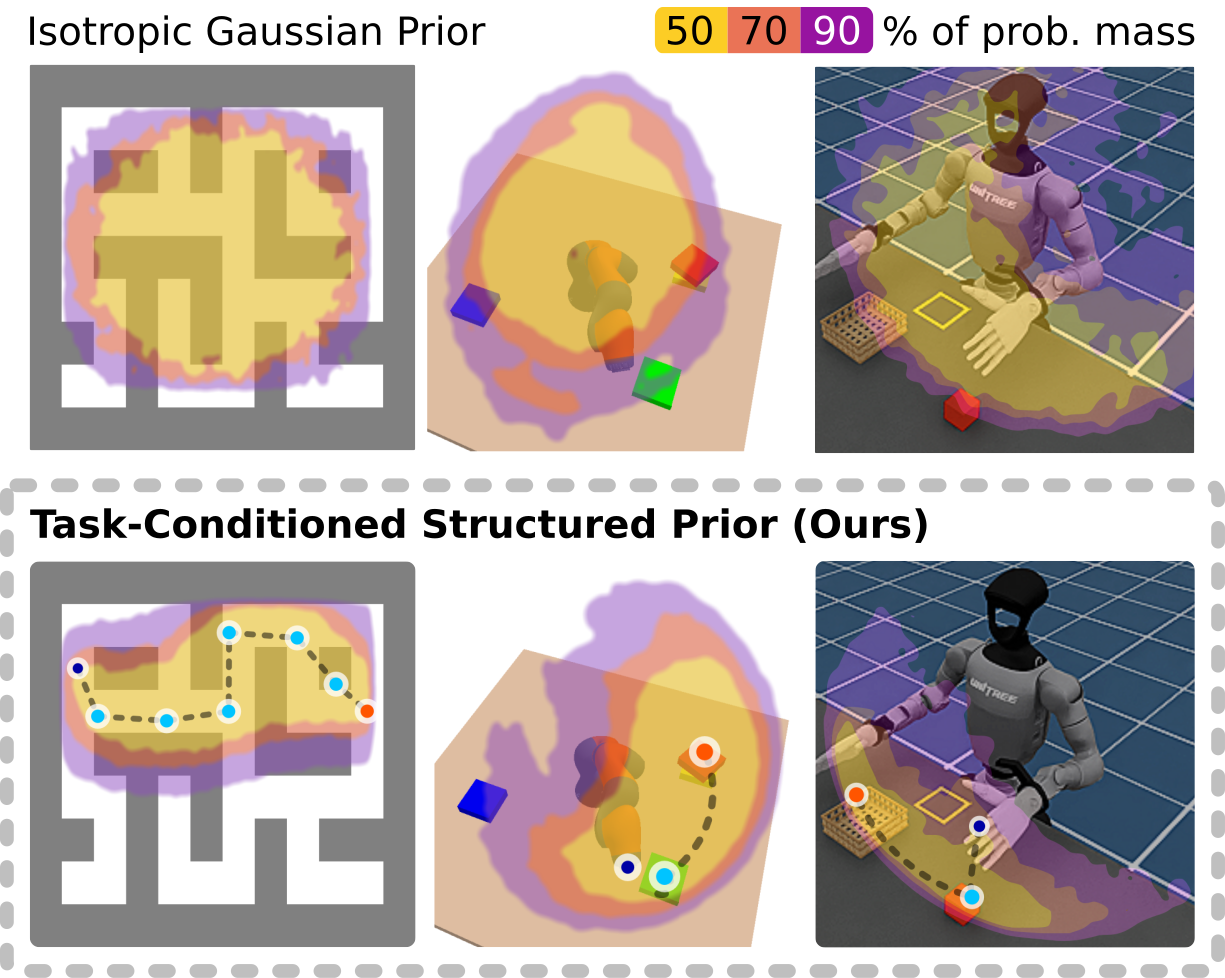}
    
\caption{
\textbf{Prior probability mass comparison
{\color{black} for Maze2D, KUKA block stacking, and G1 object-in-box tasks.}
}
The yellow, orange, and purple regions contain the densest 50\%, 70\%, and 90\% of the sampled probability mass, respectively. 
Isotropic Gaussian prior (top) spreads probability mass broadly, while our structured prior (bottom) concentrates it around task-relevant trajectory regions. Sparse, potentially imperfect key states (circles) from a lightweight upper-level module instantiate the structured prior.
} 
\label{fig:overview}
\vspace{-0.05in}
\end{figure}
Previous diffusion works typically keep the corruption isotropic and inject structural constraints through conditioning or guidance during inference \cite{carvalho2023mdp,janner2022planning,pan2024model}. {\color{black} For instance, conditioning has been used to incorporate visual context from observations or to inject goal information 
\cite{chi2023diffusion,yan2025m2diffuser}. Guidance mechanisms and auxiliary objectives incorporate dynamics, task rewards, or safety specifications into the generative process \cite{seo2025presto,xiao2023safediffuser}. In addition to conditioning and guidance, task-centric representations have been leveraged to encode constraints, for example object-centric demonstrations \cite{hsu2025spot} or SE(3)-based cost fields that couple grasp and motion optimization \cite{urain2022se3diffusionfields}. Projection-based methods have also been explored, where diffusion samples are mapped into feasible sets such as dynamically admissible trajectories \cite{bouvier2025ddat} or collision-free multi-robot paths \cite{liang2025simultaneous}. 
Beyond these categories, diffusion models have also been integrated with model-based plan \cite{yin2025diverse} or with neural dynamics models \cite{wu2025neural}.}

\textcolor{black}{Structured corruptions and source distributions have been explored more extensively outside robotics, while their use in robotic generative models remains comparatively underexplored.} {\color{black}
Examples include non-Gaussian but zero-mean choices \cite{nachmani2021non}, per-pixel variance schedules without temporal or robotics priors \cite{yu2023constructing}, and non-isotropic formulations that capture joint dependencies in human pose but still assume zero-mean noise and omit temporal correlations along the horizon, limiting their use for trajectory state conditioning \cite{curreli2025nonisotropic}. Mixture priors have also been proposed for multimodality and controllability, where Gaussian mixture parameters may be predefined through clustering \cite{jia2024structured} or heuristic structure \cite{wang2025controllable}, learned end-to-end to adaptively cover target support \cite{blessing2025end}, or dynamically adjusted during inference by solver-based approaches \cite{guo2023gaussian}. 

Recent robotics works have begun to explore this design axis through structured or informative diffusion priors. Examples include coarse trajectory predictions for diffusion-based refinement \cite{zhou2025diffrefiner}, 
diffusion bridges initialized from informative prior actions \cite{ren2025prior}, and learned diagonal Gaussian source distributions over action chunks \cite{dai2026leap}. Other approaches perform diffusion in continuous function space with temporally correlated noise \cite{chang2026fsdmp}. 
{\color{black}
While these works demonstrate the benefit of structured priors, they do not jointly encode task-specific key constraints and temporal motion structure in a full-trajectory prior. In contrast, our method constructs a structured Gaussian prior over the full trajectory by conditioning an LTV-GP on uncertain key states and timings, yielding a task-aware mean and temporally coupled covariance.
}

Hierarchical diffusion has been used to separate coarse-level decisions from finer-level motion generation through subgoals, contact plans, kinematic targets, or coarse-to-fine refinement
 \cite{chen2024simple, 
 sharma2025cascaded, dong2024diffuserlite, wang2025hierarchical,ma2024hierarchical}.
}
These approaches typically pass high-level outputs as neural conditioning while retaining isotropic low-level corruption. 
Our hierarchy instead uses the upper-level outputs to reparameterize the low-level diffusion prior itself. This yields a trajectory posterior whose mean interpolates the key states and whose covariance couples time steps, so reverse diffusion operates in a Mahalanobis geometry that biases sampling to trajectories consistent between key states while modeling key-state uncertainty. This improves feasibility without post hoc constraints or rejection.

We evaluate our approach on three domains with increasing task and embodiment complexity: Maze2D goal reaching and KUKA block stacking from the Diffuser benchmarks \cite{janner2022planning}, as well as an object-in-box manipulation task with the Unitree G1 humanoid. 

Our contributions are summarized as follows:
\begin{itemize}
\item A two-level planning framework that decomposes motion generation into sparse task-structure prediction and full-trajectory generation. 
The upper level provides task-centric key states and timings, which are treated as uncertain soft observations to construct a smooth trajectory prior for the lower level.

\item A task-conditioned diffusion formulation that uses the resulting GP posterior $(\xi,\mathcal{K})$ to parameterize the corruption and reverse processes, rather than using key states only as neural conditioning. The formulation admits closed-form forward marginals and posteriors and induces a Mahalanobis training objective.

\item \textcolor{black}{Experiments across three tasks show consistent gains over vanilla Diffuser, competitive or improved performance over conditioning- and guidance-based alternatives, and improved training and data efficiency.}

\item \textcolor{black}{A real-hardware demonstration on Unitree G1, using a diffusion planner trained entirely on simulated data, with only lightweight real-world adaptation of the upper-level key-state predictor.} 

\end{itemize}

\section{Preliminary}
\label{sec:preliminary}

\subsection{DDPM for motion planning}
\label{subsec:prelim_ddpm}
Let a robot trajectory be a stacked sequence of states $\tau=[x_1;\ldots;x_H]\in\mathbb{R}^{H \cdot d}$, where $x_t\in\mathbb{R}^d$ is the robot state at time step $t$ and $H$ is the planning horizon. We denote by $\tau^{i}$ the trajectory after $i$ forward diffusion steps.

Diffusion models learn to denoise trajectories that are gradually corrupted by Gaussian noise. Under the standard DDPM corruption \cite{ho2020ddpm}, the forward process is
\begin{equation}
\label{eq:ddpm_forward}
\tau^{i}=\sqrt{\alpha_i}\,\tau^{i-1}+\sqrt{1-\alpha_i}\,\varepsilon,
\qquad
\varepsilon\sim\mathcal{N}(0,I),
\end{equation}
where $\alpha_i:=1-\beta_i$ for $i  = 1, \ldots, N$,  $\{\beta_i\}_{i=1}^{N}\subset(0,1)$ is a scalar variance schedule, and $N$ is the total number of diffusion steps. Let $\bar{\alpha}_i:=\prod_{k=1}^{i}\alpha_k$. Then the closed-form marginal is
\[
\tau^{i}\mid\tau^{0}\sim\mathcal{N} \big(\sqrt{\bar{\alpha}_i}\,\tau^{0},\ (1-\bar{\alpha}_i)I\big).
\]
The one-step reverse posterior is also Gaussian,
\begin{equation}
\label{eq:ddpm_posterior}
q(\tau^{i-1} \mid \tau^{i},\tau^{0})
=\mathcal{N} \big(\tau^{i-1};\,\hat{\mu}_i(\tau^{i},\tau^{0}),\,\hat{\beta}_i I\big),
\end{equation}
where $\hat{\mu}_i$ and $\hat{\beta}_i$ are the posterior mean and variance coefficient at the $i^{\rm th}$ diffusion step  (closed forms omitted for simplicity).

A denoising network $\mu_{\theta}(\tau^{i},i,\texttt{cond})$ is trained to predict the posterior mean with the objective
\begin{equation}
\label{eq:ddpm_loss}
\mathcal{L}(\theta)
=\mathbb{E}_{q}\left[
\big\|\hat{\mu}_i-\mu_{\theta}(\tau^{i},i,\texttt{cond})\big\|_2^{2}
\right],
\end{equation}
where \texttt{cond} may include initial and goal states, maps, waypoints, or other task variables. At test time, sampling runs the reverse Markov chain starting from $\tau^{N}\sim\mathcal{N}(0,I)$.

In Sec.~\ref{sec:method}, we extend \eqref{eq:ddpm_forward} by biasing the mean and allowing a non-isotropic covariance in the corruption process.

\begin{figure*}[t]
\centering

\includegraphics[width=1\linewidth]{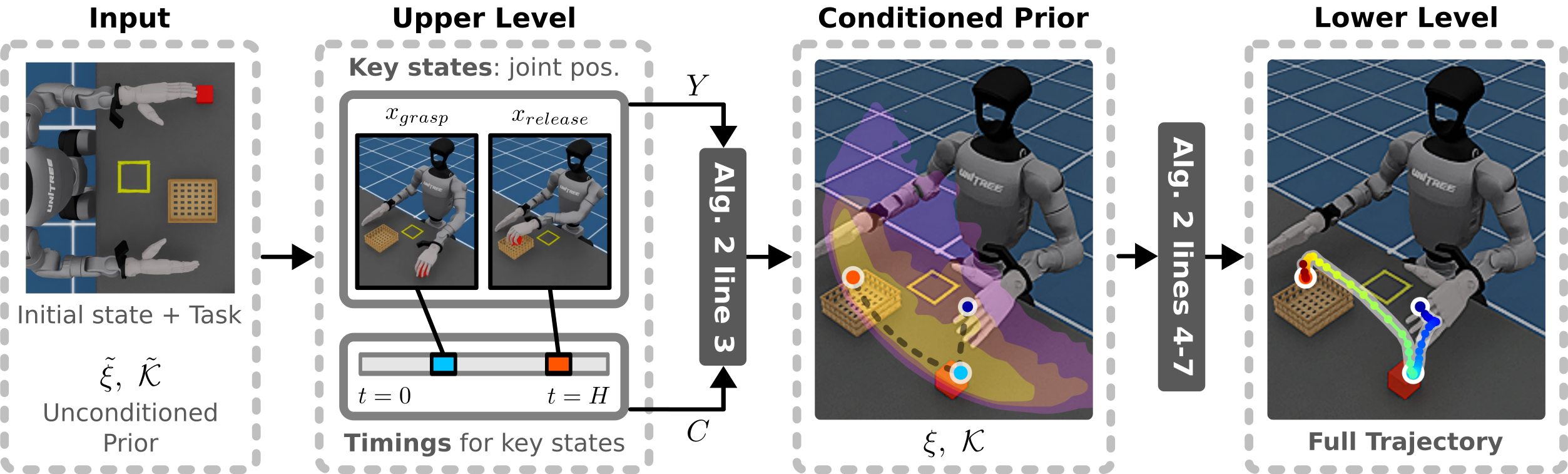}
\caption{\textbf{Trajectory generation pipeline.
}
Input: task goal and initial conditions.
Upper level produces $Y$ and $C$ for the sparse key constraints $C \cdot \tau = Y$, using task-specific learned predictors, model-based components, and/or simple rules. 
(e.g., if the constraints are simple sparse waypoints, $Y$ is a stack of key states and $C$ is a selection matrix, which corresponds to designation of timings for the key states.)
With $(Y,C)$, we construct a Gaussian prior by conditioning a GP over trajectories, yielding $(\xi,\mathcal{K})$ via \eqref{eqn:mu_goal_conditioned}--\eqref{eqn:kappa_goal_conditioned}.  
The lower level then generates the full trajectory by reverse diffusion from $\mathcal{N}(\xi,\mathcal{K})$.}
\label{fig:pipeline_overview}
\vspace{-0.1in}
\end{figure*}

\subsection{Gaussian Process Motion Planning}
\label{subsec:prelim_gpmp}
Gaussian Process Motion Planning (GPMP) models a continuous time trajectory as a Gaussian process (GP), i.e., 
\begin{equation}
\label{eq:gp_prior}
x(t)\sim\mathcal{GP}(\tilde{\xi}(t),\tilde{\mathcal{K}}(t,t')),
\end{equation}
whose mean and kernel are induced by a linear time varying (LTV) dynamics prior \cite{mukadam2018continuous}. Consider the stochastic LTV model \(\dot{x}(t)=A(t)x(t)+u(t)+w(t)\) where \(u(t)\) is control input and \(w(t)\) is zero mean white noise with spectral density \(Q_C\succeq 0\). Let the initial mean and covariance at time \(t_0\) be \(\xi_0\) and \(\mathcal{K}_0\), and let \(\Phi(t,s)\) denote the state transition matrix of \(A(\cdot)\). Then the GP has mean \(\tilde{\xi}(t)=\Phi(t,t_0)\xi_0+\int_{t_0}^{t}\Phi(t,s)u(s)\,ds\) and kernel \(\tilde{\mathcal{K}}(t,t')=\Phi(t,t_0)\mathcal{K}_0\Phi(t',t_0)^{\top}+\int_{t_0}^{\min\{t,t'\}}\Phi(t,s)Q_C\Phi(t',s)^{\top}ds\), which admit sparse and exactly computable discrete counterparts.

Following \cite{carvalho2023mdp}, our work uses a discrete-time version of (\ref{eq:gp_prior}) derived from a constant velocity or constant acceleration model with step \(\Delta t\). Writing the state at timestep \(t\) as \(x_t\), the one-step evolution is \(x_{t+1}=\Phi_{t,t+1}x_t+q_t\) with \(q_t\sim\mathcal{N}(0,Q_{t,t+1})\), where \(\Phi_{t,t+1}=\begin{bmatrix} I & \Delta t\,I\\ 0 & I \end{bmatrix}\) and \(Q_{t,t+1}=\begin{bmatrix} \tfrac{1}{3}\Delta t^{3}Q_C & \tfrac{1}{2}\Delta t^{2}Q_C\\ \tfrac{1}{2}\Delta t^{2}Q_C & \Delta t\,Q_C \end{bmatrix}\). The same structure applies, no matter if the state includes positions only or positions and velocities; velocities can be included explicitly or marginalized to obtain an equivalent GP prior over positions. In our experiments, Maze2D uses positions and velocities, whereas KUKA and G1 use joint positions only.

\section{Method}
\label{sec:method}
We first generalize DDPM to a structured Gaussian corruption process
with a nonzero mean and non-isotropic covariance
(Sec.~\ref{subsec:noniso_ddpm}).
We then describe how our hierarchical planner constructs the
structured prior and uses it for trajectory generation
(Secs.~\ref{subsec:upper_level}--\ref{subsec:lower_level}).

\subsection{DDPM with a Structured Gaussian Prior}
\label{subsec:noniso_ddpm}
Let $\mathcal{D}$ be a dataset of robot trajectories where each clean sample $\tau^0\in\mathcal{D}$ is associated with task specification that may admit multiple feasible solutions. 
Rather than corrupting trajectories toward an isotropic Gaussian
$\mathcal{N}(0,I)$, we formulate diffusion toward a task-conditioned
structured Gaussian $\mathcal{N}(\xi,\mathcal{K})$, with mean
$\xi\in\mathbb{R}^{H\cdot d}$ and covariance
$\mathcal{K}\in\mathbb{R}^{Hd\times Hd}$, where $\mathcal{K}\succ 0$. 
We first derive the corresponding forward and reverse processes for arbitrary
$(\xi,\mathcal{K})$, and then describe how our hierarchical planner instantiates
these parameters from task structure in the following subsections.

Reusing the schedule notation from Sec.~\ref{subsec:prelim_ddpm}, with $\alpha_i=1-\beta_i$ and $\bar{\alpha}_i=\prod_{k=1}^{i}\alpha_k$, we define
\begin{equation}
\label{eqn:forward_diffusion}
\tau^{i}\sim\mathcal{N}\Big(\sqrt{\alpha_i}\,\tau^{i-1}+(1-\sqrt{\alpha_i})\,\xi,\; (1-\alpha_i)\,\mathcal{K}\Big),
\end{equation}
For fixed $(\xi,\mathcal{K})$, this process can equivalently be viewed as
standard DDPM after centering by $\xi$ and whitening according to
$\mathcal{K}$. 
\eqref{eqn:forward_diffusion} yields closed form conditionals with respect to $\tau^{0}$:
\begin{equation}
\label{eq:ni_marginal}
\tau^{i}\mid\tau^{0}\sim \mathcal{N}\Big(\sqrt{\bar{\alpha}_i}\,\tau^{0}+(1-\sqrt{\bar{\alpha}_i})\,\xi,\; (1-\bar{\alpha}_i)\,\mathcal{K}\Big).
\end{equation}
The one step backward posterior is also Gaussian,
\begin{equation}
\label{eq:posterior}
q(\tau^{i-1}\mid\tau^{i},\tau^{0})=\mathcal{N}\big(\tau^{i-1};\,\tilde{\mu}_i(\tau^{i},\tau^{0},\xi),\,\tilde{\beta}_i\,\mathcal{K}\big),
\end{equation}
where 
\begin{align*}
&\tilde{\mu}_i(\tau^i,\tau^0,\xi)
:= 
\frac{\sqrt{\bar{\alpha}_{i-1}} \,\beta_i}{1-\bar{\alpha}_i}\,\tau^0
+\frac{\sqrt{\alpha_i}\,(1-\bar{\alpha}_{i-1})}{1-\bar{\alpha}_i}\,\tau^i
+ \eta_i \xi, \\[6pt]
\eta_i
&:= \frac{1}{1-\bar{\alpha}_i}
\left[
\beta_i \bigl(1 - \sqrt{\bar{\alpha}_{i-1}}\bigr)
- \sqrt{\alpha_i}\,(1 - \bar{\alpha}_{i-1})\bigl(1 - \sqrt{\alpha_i}\bigr)
\right], \\[6pt]
\tilde{\beta}_i
&= \left(\frac{\alpha_i}{1-\alpha_i} + \frac{1}{1-\bar{\alpha}_{i-1}}\right)^{-1}
\end{align*}

These are generalized forms of the standard DDPM posterior mean and variance coefficient.
(derivations are given in the Appendix). 
As $\bar{\alpha}_N\to 0$, the terminal distribution approaches
\[
\tau^{N}\sim\mathcal{N}(\xi,\mathcal{K}).
\]

In our framework, $(\xi,\mathcal{K})$ are not fixed globally, but are
instantiated for each task instance using a two-level hierarchy based on GPMP. The upper level produces sparse task-relevant key states and their timings,
which are used to condition a GPMP motion prior and obtain the
task-conditioned mean $\xi$ and covariance $\mathcal{K}$
(Sec.~\ref{sec:gpmp_prior}).
The resulting prior captures temporal smoothness while incorporating
task-specific constraints with explicit uncertainty.
The lower level then generates the full trajectory by reverse diffusion under
this structured prior.
The inputs are a task specification and an initial condition.
An overview of the framework is shown in Fig.~\ref{fig:pipeline_overview}.
The next subsections detail each level.

\subsection{Upper level: key states and timings}
\label{subsec:upper_level}

Given a full clean trajectory
$\tau=[x_1;\ldots;x_H]\in\mathbb{R}^{Hd}$,
which stacks $d$-dimensional states over a horizon of $H$,
let $Y=[y_1;\ldots;y_n]\in\mathbb{R}^{nd}$ denote a 
stack of $n$ sparse states
that are relevant for task completion, such as waypoints or states at contact events.
We refer to these as \emph{key states}.

The temporal locations of the key states are represented by a binary selection matrix
\(C\in\{0,1\}^{nd\times Hd}\), which extracts the sparse states from the full trajectory:
\[
Y=C\tau.
\]
Thus, \(C\) specifies the timings of the key states.

The upper level provides the pair $(C,Y)$. 
How $(C,Y)$ is obtained is task-dependent, as described in Sec.~\ref{sec:experiments}. 
When a learned predictor is used, it is trained with task-specific supervision on $C$, $Y$, or both.

\subsection{GPMP prior construction}
\label{sec:gpmp_prior}
\textcolor{black}{Given the upper-level output $(C,Y)$, we construct a task-conditioned
trajectory prior by conditioning the unconditioned GPMP prior introduced in
Sec.~\ref{subsec:prelim_gpmp}.
After discretization over the planning horizon, this prior is
\(
\tau \sim \mathcal{N}(\tilde{\xi},\tilde{\mathcal{K}}),
\)
where
$\tilde{\xi} = [\tilde{\xi}_1;\ldots;\tilde{\xi}_H]
\in \mathbb{R}^{Hd}$
and
$\tilde{\mathcal{K}}
= [\tilde{\mathcal{K}}(t_i,t_j)]_{1\leq i,j\leq H}$,
with $t_i$ denoting the $i$th discretized time step.
}

We treat the predicted key states $Y$ as soft observations of the trajectory
at the timings specified by $C$:
\[
Y = C\tau + \epsilon_y,
\qquad
\epsilon_y \sim \mathcal{N}(0,\mathcal{K}_y),
\]
where $\mathcal{K}_y \succeq 0$ represents the uncertainty of the key states.

Conditioning the unconditioned GPMP prior on the soft observations \((Y,C,\mathcal{K}_y)\) yields a \emph{task conditioned} posterior over the full trajectory with a closed form mean and covariance \cite{mukadam2018continuous}
\vspace{-0.05in}
\begin{align}
\xi &= \tilde{\xi} + \widetilde{\mathcal{K}}\,C^\top\left(C\,\widetilde{\mathcal{K}}\,C^\top + \mathcal{K}_y\right)^{-1}\big(Y - C\,\tilde{\xi}\big), \label{eqn:mu_goal_conditioned}\\
\mathcal{K} &= \widetilde{\mathcal{K}} - \widetilde{\mathcal{K}}\,C^\top\left(C\,\widetilde{\mathcal{K}}\,C^\top + \mathcal{K}_y\right)^{-1}C\,\widetilde{\mathcal{K}}. \label{eqn:kappa_goal_conditioned}
\end{align}

This posterior \(\mathcal{N}(\xi,\mathcal{K})\) serves as the task-conditioned prior for the lower-level diffusion model. 
Small variances in \(\mathcal{K}_y\) enforce tight adherence to key states (e.g., fixed initial or goal states), while larger variances allow flexibility (e.g., uncertain waypoints). Thus, \(Y\) anchors the trajectory near important states and \(\mathcal{K}_y\) encodes confidence in those anchors, without imposing hard constraints. 

\subsection{Lower level: trajectory generation by denoising}
\label{subsec:lower_level}
The conditioned prior $\mathcal{N}(\xi,\mathcal{K})$, constructed as described
in Secs.~\ref{subsec:upper_level}--\ref{sec:gpmp_prior}, determines the
task-conditioned noise geometry of the lower-level diffusion. 
At inference, the lower level initializes
$\tau^N\sim\mathcal{N}(\xi,\mathcal{K})$ and iteratively samples from the
learned reverse transitions to obtain the final trajectory $\tau^0$. Following the standard DDPM parameterization, we model the learned reverse
transition as
\[
p_\theta(\tau^{i-1}\mid\tau^i,\texttt{cond})
=
\mathcal{N}\!\left(
\tau^{i-1};
\mu_\theta(\tau^i,i,\texttt{cond}),
\tilde{\beta}_i\mathcal{K}
\right),
\]
where $\texttt{cond}$ denotes the task information provided to the denoising
network.
We fix the reverse covariance to the analytical posterior covariance
$\tilde{\beta}_i\mathcal{K}$ in \eqref{eq:posterior} and learn the posterior
mean with $\mu_\theta$.
The corresponding variational objective~\cite{ho2020ddpm} minimizes the 
KL divergence between the true and learned reverse distributions.
Because both use the covariance $\tilde{\beta}_i\mathcal{K}$, their
KL divergence reduces to
\begin{equation}
D_{\mathrm{KL}}\!\left(
q
\,\|\, 
p_\theta
\right)
=
\frac{1}{2\tilde{\beta}_i}
\big\|
\tilde{\mu}_i-\mu_\theta(\tau^i,i,\texttt{cond})
\big\|_{\mathcal{K}^{-1}}^2,
\end{equation}
up to terms independent of $\theta$, where
$\|v\|_A^2:=v^\top A v$.
Thus, omitting the scalar schedule-dependent factor, we train with the
Mahalanobis MSE
\begin{equation}
\label{eq:mah_loss}
L(\theta)
=
\mathbb{E}_{q}\left[
\big\|
\tilde{\mu}_i-\mu_\theta(\tau^i,i,\texttt{cond})
\big\|_{\mathcal{K}^{-1}}^2
\right].
\end{equation}
Because $\mathcal{K}$ is the covariance of the conditioned GP prior,
this objective weights prediction errors according to its task-dependent
uncertainty and temporal correlation structure.

During lower-level training, we use the ground-truth key states and timings
$(C,Y)$ as the nominal upper-level outputs and perturb them to account for
prediction errors at test time.
We perturb the key-state timings encoded by $C$ and reconstruct the
corresponding selection matrix $C_{\mathrm{pert}}$, while perturbing the key
states as
\(
Y_{\mathrm{pert}} = Y + dY_\omega,
\)
where $dY_\omega$ denotes a sampled key-state perturbation.
The perturbed pair $(C_{\mathrm{pert}},Y_{\mathrm{pert}})$ is then used to
instantiate the task-conditioned prior $(\xi,\mathcal{K})$ via
\eqref{eqn:mu_goal_conditioned}--\eqref{eqn:kappa_goal_conditioned}.
We then update the lower-level diffusion parameters using the Mahalanobis MSE loss \eqref{eq:mah_loss}, as summarized in Alg.~\ref{alg:train_lower}. The complete inference process, including both the upper and lower levels, is summarized in Alg.~\ref{alg:test}.

\begin{remark}
Computing \textcolor{black}{covariance-related terms} \(\big(C\widetilde{\mathcal{K}}C^\top + \mathcal{K}_y\big)^{-1}\), \(\mathcal{K}\), and \(\mathcal{K}^{-1}\) repeatedly can be costly. 
We therefore precompute the corresponding gain terms for all timing configurations used at training and inference. 
For example, in the KUKA manipulation task, we divide the horizon \(H\) into 10 bins and precompute gains for all grasp and release bin pairs \(\tbinom{10}{2}=45\), since only \(C\) changes with timing choices. 
This preserves time range conditioning while keeping online costs low.
\end{remark}

\begin{algorithm}[t]
\caption{Lower-level training (for full trajectory)}
\label{alg:train_lower}
\begin{algorithmic}[1]
\State \textbf{Inputs:} dataset $\mathcal{D}$, GP prior $(\tilde{\xi},\widetilde{\mathcal{K}})$
\State Initialize lower-level  diffusion module parameters $\theta$
\While{not converged}
  \State \textcolor{black}{Sample $\tau^0 \sim \mathcal{D}$ and extract $(C,Y)$ from $\tau^0$}  
  \State \textcolor{black}{Perturb the key-state timings encoded by $C$ and reconstruct $C_{\mathrm{pert}}$; $Y_{\mathrm{pert}} \gets Y + dY_\omega$}
\State $(\xi,\mathcal{K}) \gets \text{GP-Post}(\tilde{\xi},\widetilde{\mathcal{K}}, C_{\mathrm{pert}}, Y_{\mathrm{pert}}, \mathcal{K}_y)$ via \eqref{eqn:mu_goal_conditioned}--\eqref{eqn:kappa_goal_conditioned}
  \State Sample $i \sim \text{Unif}\{1,\dots,N\}$
  \State $\tau^i \sim \mathcal{N}\left(
     \sqrt{\bar\alpha_i}\tau^0 + (1-\sqrt{\bar\alpha_i})\xi,\;
     (1-\bar\alpha_i)\mathcal{K}\right)$
  \State Gradient descent $\nabla_\theta \left\| \tilde{\mu}_i(\tau^i,\tau^0,\xi) - \mu_\theta(\tau^i,i) \right\|_{\mathcal{K}^{-1}}^2$

\EndWhile
\end{algorithmic}
\end{algorithm}

\begin{algorithm}[t]
\caption{Test-time Planning (Inference)}
\label{alg:test}
\begin{algorithmic}[1]
\State \textbf{Inputs:} task specification, unconditioned GP prior $(\tilde{\xi},\widetilde{\mathcal{K}})$
\State \textbf{Upper level:} produce $(C, Y)$  \Comment{using the task-specific key-state/timing predictor or heuristic}
\State 
$(\xi,\mathcal{K}) \gets \text{GP-Post}(\tilde{\xi},\widetilde{\mathcal{K}}, C, Y, \mathcal{K}_y)$ via \eqref{eqn:mu_goal_conditioned}--\eqref{eqn:kappa_goal_conditioned}
\State \textbf{Lower Level:} 
\State $\tau^N \sim \mathcal{N}(\xi,\mathcal{K})$
\For{$i = N,  \dots, 1$}
  \State $\tau^{i-1} \sim \mathcal{N}( \mu_{\theta}(\tau^i,i),\tilde \beta_i\mathcal{K}) $
\EndFor
\State \textbf{return} $\tau^0$
\end{algorithmic}
\end{algorithm}

\begin{remark}
Motion Planning Diffusion (MPD) \cite{carvalho2023mdp} applies a GP as a guidance cost only during reverse sampling, while the forward corruption remains isotropic. We instead parameterize the noise with a GP: conditioning on key states sets the mean \(\xi\) and covariance \(\mathcal{K}\) that govern both the forward and posterior distributions.
\end{remark}

\begin{table}[t]
\centering
\begin{tabular}{l c c c}
\toprule
\textbf{Method \textbackslash Task} & \textbf{Maze2D} & \textbf{KUKA}  & \textbf{G1} \\
\midrule
Diffuser & 51.6 $\pm$ 6.2  & 23.6 $\pm$ 4.7 & 53.8 $\pm$ 6.3 \\
Diffuser + KeyCond & 59.4 $\pm$ 1.8  & 55.8 $\pm$ 4.2 & \textbf{69.2 $\pm$ 4.3} \\
EDMP & \textbf{78.4  $\pm$ 2.9 } 
& 30.8 $\pm$ 4.5 & 59.0 $\pm$ 1.4  \\
\textbf{Structured-Diffuser} & \textbf{74.4  $\pm$ 2.6 } & \textbf{67.2 $\pm$ 2.5 }  & \textbf{69.6 $\pm$ 2.8 }\\
\bottomrule
\end{tabular}
\caption{Task success rate (\%) in simulation (mean \(\pm\) standard deviation over five seeds, with 100 trials per seed). 
}
\label{tab:success_rates}
\vspace{-0.025in}
\end{table}

\begin{figure}[t]
\centering
\includegraphics[width=1\linewidth]{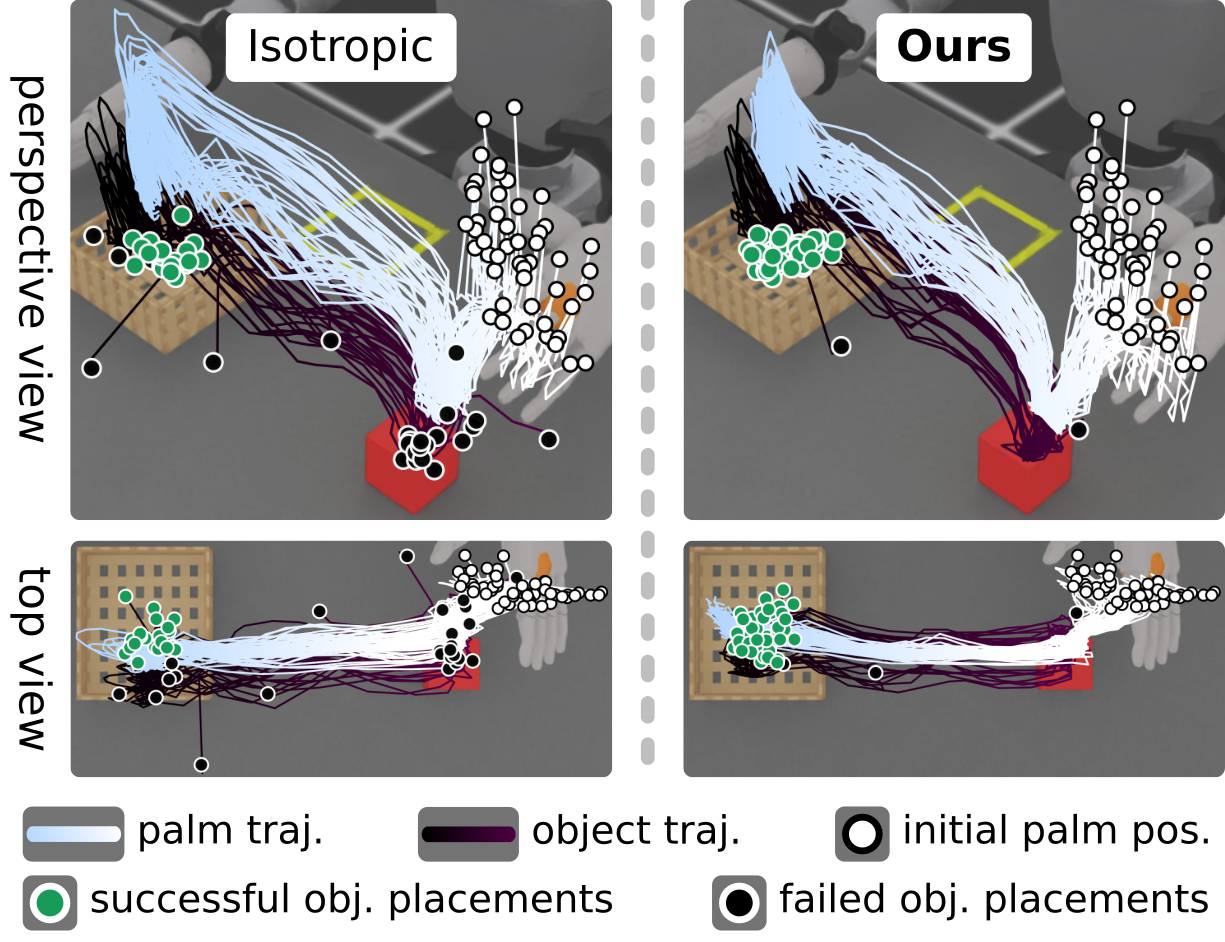}

\caption{\textcolor{black}{
\textbf{Distribution of end-effector trajectories for the G1 object-in-box task.}
For a representative object–box configuration, 50 sampled plans are executed in Isaac Sim. 
Our method (right) concentrates trajectories around appropriate grasp and release regions, resulting in fewer failures.
}}
\label{fig:traj_distriution_g1}
\vspace{-0.1in}
\end{figure}

\begin{figure}[t]
\centering
    \includegraphics[width=0.9\linewidth,
    trim=0cm 0.05cm 0cm  0.2cm,
    clip]{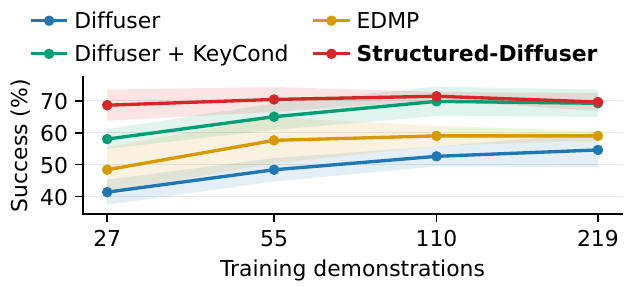}
\caption{
\textbf{Data efficiency for G1 Put-In-Box.} Success rate is plotted over number of training demonstrations.
Points are means over five evaluation seeds of 100 episodes each and shading shows
standard deviation. 
}
\label{fig:data_efficiency}
\vspace{-0.025in}
\end{figure}

\begin{figure}[t]
\centering
\includegraphics[width=1\linewidth, trim={0.13cm 0.1cm 0.35cm 0}, clip]{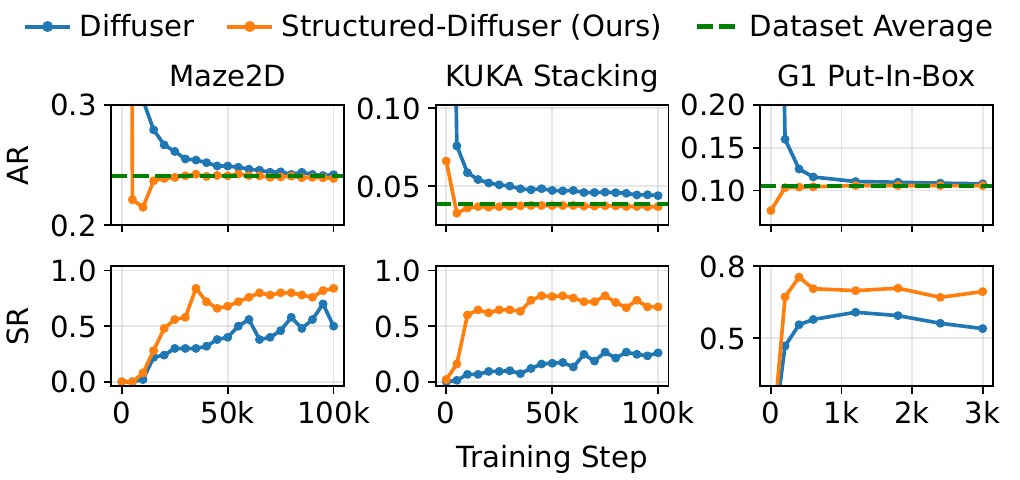}
\caption{\textbf{Average roughness (AR) and success rate (SR) 
over training for three tasks.} Metric calculated as average of 50 samples. 
The dashed line shows the dataset-average AR as a reference for reasonable smoothness.}
\vspace{-0.1in}
\label{fig:success_smooth_compare}
\end{figure}

\begin{figure}[t]
\centering
\includegraphics[width=1\linewidth]{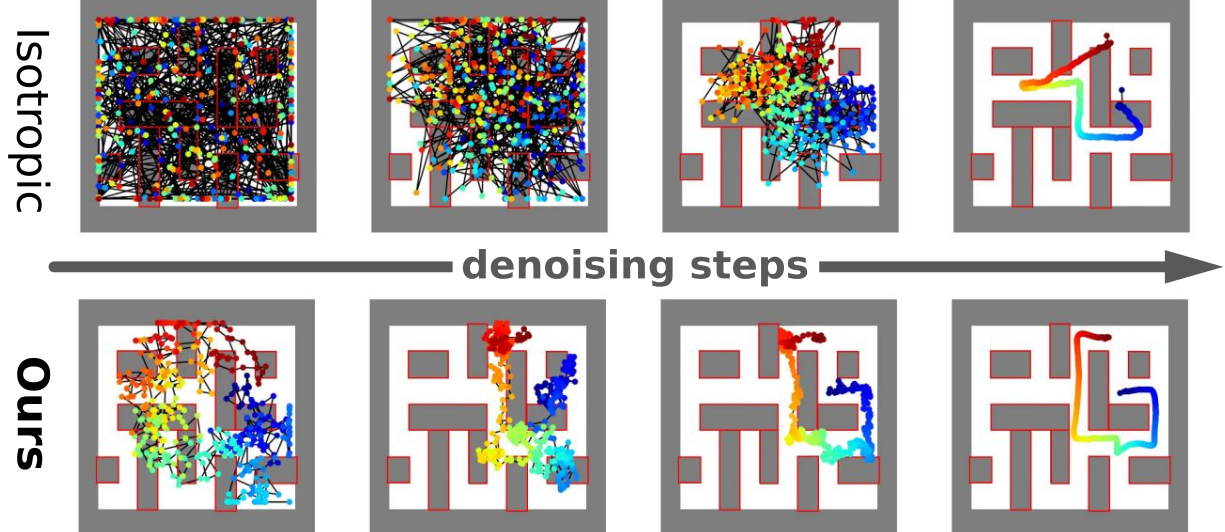}

\caption{\textbf{Reverse diffusion in Maze2D.}
Intermediate denoising steps from left to right.
Our prior rapidly converges to smooth, goal-reaching trajectories.}
\label{fig:denosing_process_comparison}
\vspace{-0.05in}
\end{figure}

\begin{figure}[t]
\centering
\includegraphics[width=1\linewidth]{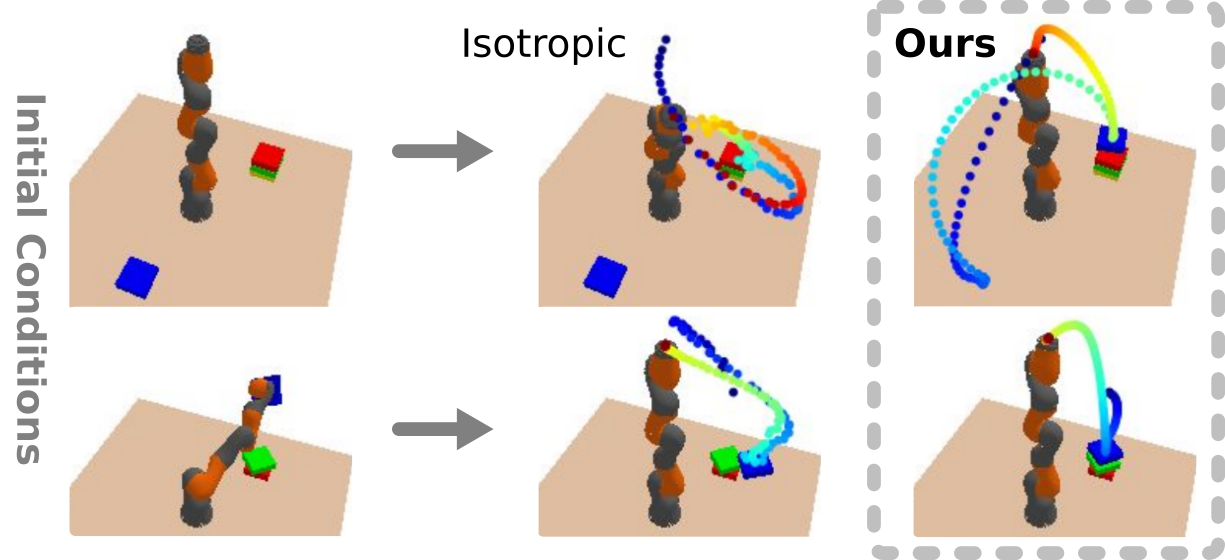}
\caption{\textbf{Comparison of generated trajectories for KUKA block stacking.} 
Each row shows a task with a random initial arm configuration and block positions. 
Columns: (left) initial condition, (middle)  trajectories generated with an isotropic Gaussian prior, (right) trajectories generated with our model.}
\vspace{-0.1in}
\label{fig:traj_visualization_kuka}
\end{figure}

\section{Experiments}

\label{sec:experiments}
\subsection{Experimental Setup}
We evaluate the proposed structured-noise formulation on three domains. 
We describe the benchmark tasks and how key states and timings are defined.

\textbf{(i) Maze2D goal reaching. }
Each task specifies a start and goal state on a fixed maze. Key states are defined as evenly spaced temporal waypoints between them, so \(C\) is fixed by design because the waypoint timings are predetermined.
The upper-level key-state predictor is a lightweight diffusion model trained to generate the waypoint set \(Y\) at the fixed timings \(C\).

\textbf{(ii) KUKA arm block stacking. }
The task is to grasp and stack blocks with a 7-DoF KUKA arm.
The Diffuser benchmark provides joint-position trajectories, per-block binary contact flags, and block-position trajectories.
Since block positions are determined from kinematics and contact, we train the models only on joint trajectories and contact flags.
The key-state stack \(Y\) contains the initial, grasp, and release joint configurations, while their corresponding indices define \(C\).
A lightweight upper-level diffusion model predicts the contact-flag trajectory, from which grasp and release timings are extracted to construct \(C\).
For \(Y\), we use the grasp and release joint configurations from the reference trajectory for each evaluation scene. Thus only the timings are predicted in this task, and Diffuser + KeyCond receives the same \(Y\).

At test time, the grasp target is the currently grasped block if any, or otherwise the block nearest the initial end-effector position. The release target is the highest remaining block.

\textbf{(iii) Unitree G1 object-in-box manipulation.}
The G1 humanoid grasps an object and places it inside a target box.
The trajectory state consists of 7-DoF left-arm joint positions and a hand grasping fraction.
The key-state stack  \(Y\) contains the initial, grasp, and release configurations.
We train an MLP-based inverse-kinematics (IK) predictor for the grasp and release joint configurations, and another MLP to predict their timings for constructing \(C\).
The dataset consists of 219 successful \textcolor{black}{Finite State Machine}-generated trajectories collected in Isaac Sim.

Simulation evaluations randomize start--goal pairs for Maze2D, initial joint configurations and block layouts for KUKA, and object/box poses and initial arm configurations for G1.

\subsection{Baselines and Ablations}
We compare three planners in addition to ours that share the same UNet backbone and training schedule, differing only in how task structure is incorporated. 
All models are trained for the same number of optimization steps (Tasks from Diffuser benchmark -- 100k steps with batch size 32, G1 -- 3k steps with batch size 512)
The detailed design of each planner is as follows:

\textbf{(i) Diffuser.} 
    The baseline Diffuser~\cite{janner2022planning}, which uses isotropic Gaussian noise and conditions only on task inputs such as initial and goal states.
    
 \textbf{(ii) Diffuser + KeyCond.} 
    A hierarchical variant of Diffuser where the lower-level diffusion is conditioned on key states and their timings produced by the upper level, while still using isotropic Gaussian noise. This isolates the effect of neural task conditioning without structured noise.
       
\textbf{(iii) EDMP~\cite{saha2024edmp}.} 
A single-level Diffuser guided at inference time by model-based costs. 
Following the original formulation, we use task-specific costs for each benchmark. 
For Maze2D, the cost penalizes obstacle penetration at each state and at midpoints between adjacent states to approximate swept-volume collisions. This implementation requires access to the maze geometry at test time. 
For KUKA and G1 manipulation, since~\cite{saha2024edmp} does not specify a cost for grasp-and-place tasks, we design a simple heuristic cost based on the minimum distance between the end-effector and the target, rewarding configurations with a small distance. 

\textbf{(iv) Structured-Diffuser (Ours).}
    Our two-level approach, where the upper level provides key states and timings, and the lower level denoises under a GPMP-conditioned prior.

\textcolor{black}{
Note that the choice of noise prior applies only to the lower-level diffusion.  
Whenever the upper level uses diffusion, it employs a standard DDPM with isotropic Gaussian.
}

\subsection{Results and Analysis}

\textbf{Task success and data efficiency.} In Maze2D, success requires collision-free goal reaching; in KUKA block stacking, a successful final block stack, verified by kinematic replay of the generated trajectory; and in G1 object-in-box, successfully grasping the object and placing it inside the target box in physics-based simulation (Isaac Sim). 
These binary criteria differ from the metrics in \cite{janner2022planning}, so absolute numbers are not comparable.
Table~\ref{tab:success_rates} shows that Structured-Diffuser outperforms vanilla Diffuser across all three tasks. 
We attribute these gains to the structured prior, which concentrates probability mass around task-critical regions through the soft GP conditioning in
\eqref{eqn:mu_goal_conditioned}--\eqref{eqn:kappa_goal_conditioned}.
This effect is visualized in Fig.~\ref{fig:traj_distriution_g1}, where sampled G1 trajectories concentrate near the object during grasping and near the box during release.

Among all methods, Structured-Diffuser achieves the highest mean success rate on KUKA and G1.
On Maze2D, Structured-Diffuser achieves the second-highest success rate (74.4\%), behind EDMP (78.4\%). However, EDMP uses ground-truth map geometry at test time, whereas Structured-Diffuser does not.
Diffuser + KeyCond performs similarly to Structured-Diffuser on full-data G1.
With less training data, however, the gap widens: using 27 demonstrations, Structured-Diffuser retains 68.6\% compared with 58.0\% for KeyCond (Fig.~\ref{fig:data_efficiency}). 

\textbf{Smoothness and inter-state consistency.} 
We evaluate trajectory smoothness using the \emph{Average Roughness} (AR) metric~\cite{saha2024edmp}, defined as the average of $\|x_{t+1}-x_t\|_2$ along the trajectory.
Because excessively low AR can correspond to little or no motion, we compare generated trajectories with the dataset-average AR in Fig.~\ref{fig:success_smooth_compare}.
Representative trajectories and denoising processes are shown in Figs.~\ref{fig:denosing_process_comparison} and~\ref{fig:traj_visualization_kuka}.
These quantitative and qualitative results show that Structured-Diffuser reaches dataset-level smoothness earlier and generates smooth trajectories more consistently. This behavior is encouraged by LTV-GP-based covariance, which encodes temporal dependencies across trajectory states.

\begin{figure}[t]
\centering
    \includegraphics[width=1\linewidth
    ]{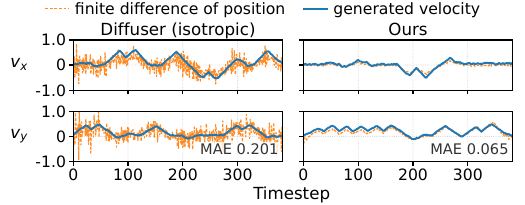}
    
\caption{\textbf{Maze2D velocity consistency.} 
Left: isotropic Gaussian prior; Right: our model. 
Orange: numerical derivatives of generated positions; Blue: generated velocities. 
Closer alignment indicates better inter-state (pos \& vel) consistency.
}
\label{fig:dyn_feas_maze2d}
\end{figure}

In Maze2D, the LTV-GP-based covariance explicitly couples position and velocity across the trajectory.
We evaluate this by comparing generated velocities with finite-difference derivatives of generated positions (Fig.~\ref{fig:dyn_feas_maze2d}).
Structured-Diffuser achieves substantially lower velocity MAE than the baselines (Table~\ref{tab:maze2d_vel_mae}), demonstrating stronger position--velocity consistency.

\textbf{Training efficiency.} 
As shown in Fig.~\ref{fig:success_smooth_compare}, Structured-Diffuser reaches dataset-level smoothness and high success rates earlier in training than vanilla Diffuser.
This suggests that encoding task and motion structure in the prior reduces the amount of training needed to obtain useful trajectories.

\textbf{Phase-dependent task structure.}
\textcolor{black}{
In many robotics tasks, desired behavior depends not only on the robot state but also on when it occurs (e.g., contact timing). This timing--configuration coupling is difficult to encode with state-wise guidance, especially in partially denoised trajectories where timings may shift under noise. Our framework instead predicts task-phase timings at the upper level and conditions the lower-level prior on them with uncertainty, which is useful for  manipulation tasks with critical interaction timing.
}

\begin{table}[t]
\centering
{\color{black}

\begin{tabular}{l m{1.25cm} m{1.25cm} m{1.25cm} m{1.25cm}}
\toprule
\textbf{Method}
& \textbf{Diffuser}
& \textbf{Diffuser + KeyCond}
& \textbf{EDMP}
& \textbf{Structured-Diffuser} \\
\midrule
\textbf{Vel MAE}
& 0.2187
& 0.2064
& 0.2012
& \textbf{0.0767} \\
\bottomrule
\end{tabular}

\caption{
Maze2D velocity consistency: mean absolute error (MAE) between generated velocities and finite-difference derivatives of positions over 100 trajectories.
}
\label{tab:maze2d_vel_mae}
}
\vspace{-0.1in}
\end{table}

\begin{figure}[t]
\centering
\includegraphics[width=1\linewidth]{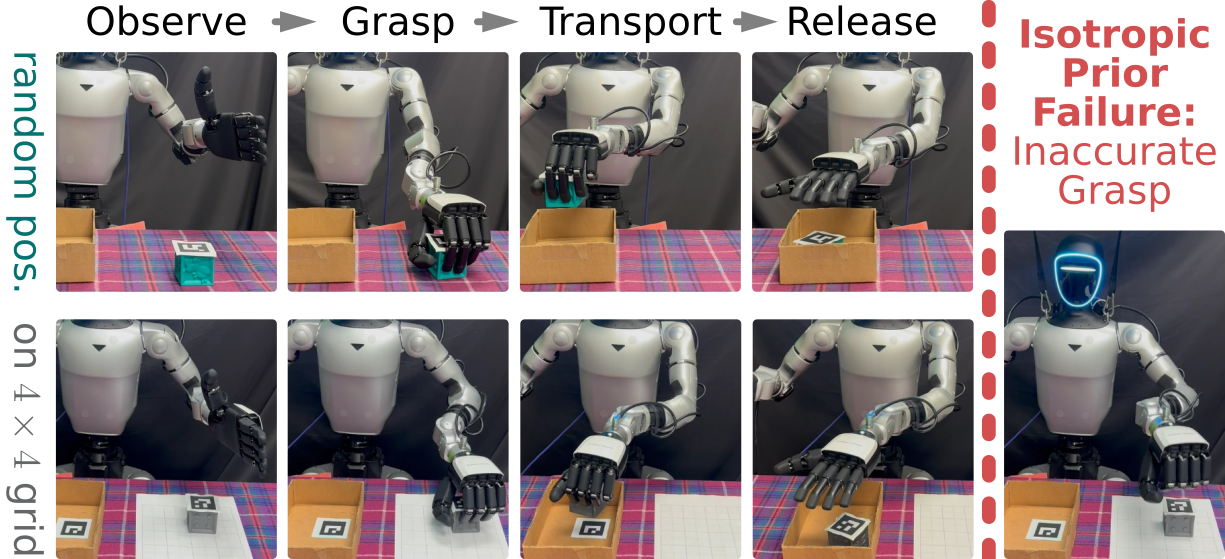}
\caption{\textcolor{black}{\textbf{Representative G1 hardware rollouts.} The robot localizes the object with the mounted camera and executes pick-and-place into the box. Top: randomly placed object. 
Bottom: object placed on a fixed \(4\times4\) grid, identical across planners.
Right: isotropic-prior failure due to an inaccurate grasp at the same grid location shown in the bottom sequence. 
}}
\vspace{-0.1in}
\label{fig:hardware}
\end{figure}

\subsection{Hardware Experiment}
\label{subsec:hardware_exp}

We deploy the planner on a physical Unitree G1 humanoid equipped with an Inspire hand, with the torso held at a fixed reference \textcolor{black}{by a low-level locomotion controller.}
ArUco markers provide object and box poses.
The planner generates a 60-step trajectory, which is interpolated and executed at 20 Hz without online replanning. 
We first evaluate randomized-location trials using models trained on the full simulation dataset, with object and box positions sampled within \(15\,\mathrm{cm}\times15\,\mathrm{cm}\) and \(7\,\mathrm{cm}\times7\,\mathrm{cm}\) regions, respectively.
Representative rollouts are shown in Fig.~\ref{fig:hardware}. 

\textcolor{black}{
For quantitative evaluation, we test 16 designated object locations on a \(4\times4\) grid using diffusion models trained on a separate simulation dataset of 63 trajectories, whose size is deliberately reduced to demonstrate data efficiency.
For sim-to-real transfer, we collect 19 grasp configurations (rather than full trajectories) in real hardware, 
}
and fine-tune only the key-state predictor on those configurations for less than one minute; the diffusion planners remain fixed.

Each planner was evaluated once at each of the 16 grid locations. Structured-Diffuser succeeded in 8/16 trials, compared with 3/16 for Diffuser.

\section{Conclusion}
In this paper, we propose a hierarchical diffusion planner that embeds task and motion structure directly in the noise model.
We validate the approach 
\textcolor{black}{across three robotic domains, demonstrating improved task success, trainig efficiency and, where evaluated, data efficiency and inter-state consistency. We further demonstrate hardware deployment on a physical G1 humanoid. }
\textcolor{black}{The lower-level structured diffusion is agnostic to how the key states and timings are obtained; in our experiments, the upper level combines diffusion models, lightweight MLPs, inverse kinematics, and simple task-specific rules. }
Moreover, our method does not require handcrafted reward terms or auxiliary objectives for guidance, and it does not rely on advanced neural architectures beyond conditioning on upper-level-generated features. 

Future work includes disentangling the contributions of the prior mean and covariance, extending  the framework to more complex whole-body humanoid manipulation, investigating richer upper-level task specifiers, such as vision-language models,  
\textcolor{black}{and more systematically studying sensitivity to upper-level module variants and related hyperparameters.}

\section{Acknowledgment}
ChatGPT and Claude were used in
polishing the writing. 
CODEX and Claude Code were used to polish our code base.

\bibliographystyle{IEEEtran}
\bibliography{bibtex/bib/IEEEexample}

\appendix

First, we drive the marginal $q(\tau^j \mid \tau^0)$ \eqref{eq:ni_marginal} by induction. 
Assume \eqref{eq:ni_marginal} holds for \(i=j-1\). Then, \(\tau^j \mid \tau^0\) is Gaussian with mean and variance
\begin{align*}
&\mathbb{E}[\tau^j \mid \tau^0]
= \sqrt{\alpha_j}\,\mathbb{E}[\tau^{j-1} \mid \tau^0] + (1 - \sqrt{\alpha_j})\xi \\
&= \sqrt{\alpha_j}\left(\sqrt{\bar{\alpha}_{j-1}} \tau^0 + (1 - \sqrt{\bar{\alpha}_{j-1}})\xi\right) + (1 - \sqrt{\alpha_j})\xi \\
&= \sqrt{\bar{\alpha}_j}\,\tau^0 + (1 - \sqrt{\bar{\alpha}_j})\xi,\\
&\mathrm{Var}[\tau^j \mid \tau^0]
= \alpha_j (1 - \bar{\alpha}_{j-1})\mathcal{K} + (1 - \alpha_j)\mathcal{K} = (1 - \bar{\alpha}_j)\mathcal{K}.
\end{align*}
Hence the form holds for step \(j\), as claimed.

Second, we derive the posterior \eqref{eq:posterior}.
Both $q(\tau^i \mid \tau^{i-1})$ and $q(\tau^{i-1} \mid \tau^0)$ are Gaussian distributions with known forms.
Using Bayes' rule on
$
q(\tau^{i-1} \mid \tau^i, \tau^0)
\propto q(\tau^i \mid \tau^{i-1}) \; q(\tau^{i-1} \mid \tau^0)$,   we obtain \eqref{eq:posterior}.

\end{document}